\documentclass{article}

\usepackage[preprint]{neurips_2019}

\usepackage[]{neurips_2019}

\usepackage[utf8]{inputenc} 
\usepackage[T1]{fontenc}    
\usepackage{hyperref}       
\usepackage{url}            
\usepackage{booktabs}       

\usepackage{amsfonts}       
\usepackage{nicefrac}       

\usepackage{microtype}      
\usepackage{graphicx}
\graphicspath{ {images/} }
\usepackage{subfig}

\title{Divide and Conquer: an Accurate Machine Learning Algorithm to Process Split Videos on a Parallel Processing Infrastructure}

\author{%
   Walter Mayor Toro\\
  Department of Automatics and Electronics\\
  Universidad Autónoma de Occidente\\
  \texttt{wmmayor@uao.edu.co} \\
  \And
  Juan C. Perafan Villota\\
  Department of Automatics and Electronics\\
  Universidad Autónoma de Occidente\\
  \texttt{jcperafan@uao.edu.co} \\
  \And
     Oscar H. Mondragon \\
 Department of Automatics and Electronics \\
  Universidad Autónoma de Occidente\\
  \texttt{ohmondragon@uao.edu.co } \\
   \And
      Johan S. Obando-Ceron\\
  Department of Automatics and Electronics\\
  Universidad Autónoma de Occidente\\
  \texttt{jsobando@uao.edu.co} \\
}

\begin{document}

\maketitle

\begin{abstract}
Every day the number of traffic cameras in cities rapidly increase and huge amount of video data are generated. Parallel processing infrastructure, such as Hadoop, and programming models, such as MapReduce, are being used to promptly process that amount of data. The common approach for video processing by using Hadoop MapReduce is to process an entire video on only one node, however, in order to avoid parallelization problems, such as load imbalance, we propose to process videos by splitting it into equal parts and processing each resulting chunk on a different node. We used some machine learning techniques to detect and track the vehicles. However,  video division may produce inaccurate results. To solve this problem we proposed a heuristic algorithm to avoid process a vehicle in more than one chunk.

\end{abstract}

\section{Research problem and motivation}
According to [1], by 2020 one billion cameras will be installed throughout cities. More cameras implicate a generation of huge amount of video data.
Fortunately, parallel processing ecosystem based on distributed environments have permitted to face  computational complexity and big datasets. Few intelligent traffic management systems  (ITMS) approaches have been developed using parallel processing infrastructure. Some of those solutions have used the Hadoop MapReduce ecosystem which allows parallelization of multiple jobs by using several computers or multiple processors in one computer[2]. 

In ITMS, researches have used Hadoop MapReduce to sequentially analyze multiple videos by using batch processing, where each entire video file is only processed by a unique node [3][4]. In this case, when the videos are reasonably long, it can arise a problem known as load imbalance. On the other hand, this way of processing videos only allows to take the maximum advantage of the parallel ecosystem when the dataset contains a big number of videos [5]. To avoid these problems we have divided each video to process them by chunks in  different nodes. Unfortunately, this approach brings some problems. For example, if we divide a video, we could find a vehicle in different video chunks,  possibly affecting the precision of the analysis algorithm. For the case of a counting vehicles system, classical algorithms would detect more vehicles in video chunks that when an unsplit video is analyzed.

Motivated by the idea of processing chunk videos within a parallel processing ecosystem without obtaining wrong results, we propose an intelligent video processing algorithm to process the chunks and avoiding to take the same vehicle into account many times. In the framework we propose, it does not matter the number and size of chunks in which the video is divided. Machine learning techniques are used to process videos and detect vehicles. The considerations above motivate the formulation of the following research question:  How to process chunks of video using Hadoop MapReduce in order to decrease the computational time and reach accurate results?

\section{Technical contribution}
\begin{figure}[h]
  \centering
  \includegraphics[width=0.7\textwidth]{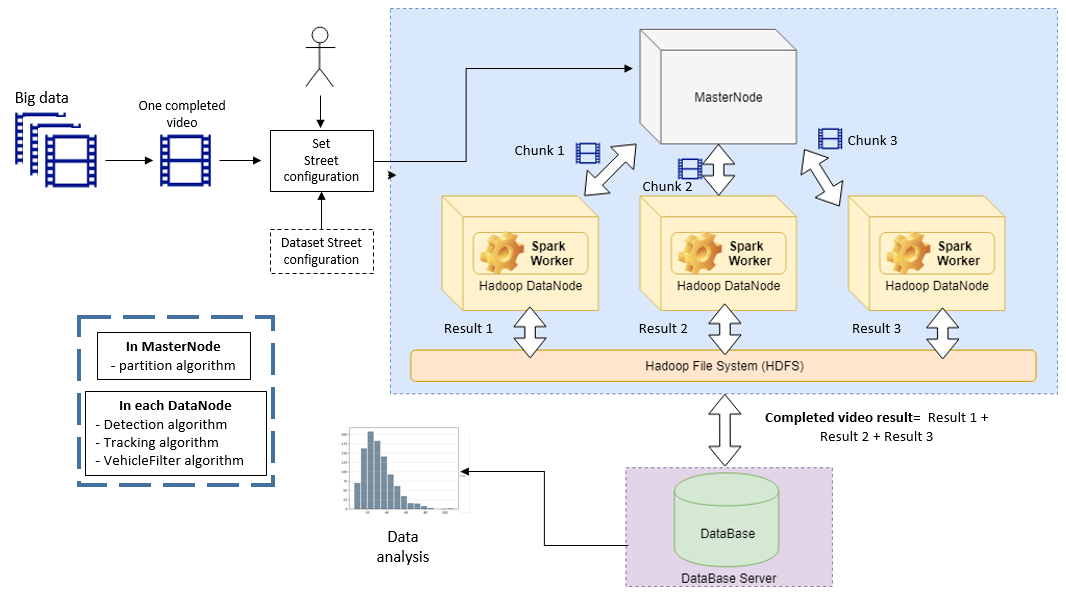}
  \caption{Framework proposed.}
  \label{fig:framework}
\end{figure}

Our proposed algorithm takes as input datasets of videos from traffic cameras. Some of the videos we used were taken from YouTube and other videos from traffic cameras of Cali (Colombia) city. We developed a software platform that allows a user to select a street from a video, where the vehicle counting will be performed. By drawing lines, the user is able to specify the target counting area. This configuration is saved in a dataset to be used in futures videos. In the master node, a partitioning algorithm divides the video in chunks, taking into account the number of data nodes. Each chunk video is sent to a Hadoop data node, where 3 algorithms are applied: first, we detect the vehicles in the chunk by using the deep learning algorithm YOLO; then, we track each vehicle detected by using kalman filters and the hungarian algorithm; and finally, we filter the vehicles that were registered in others chunk videos. The videos are analyzed and the result is saved in a database than can be continuously accessed for probabilistic analysis in vehicular traffic tasks (see Figure \ref{fig:framework}).   

\begin{figure*}[ht!]
   \subfloat[\label{genworkflow}]{%
      \includegraphics[  width=0.16\textwidth]{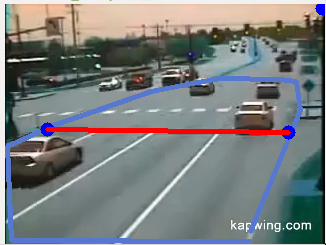}}
   \subfloat[\label{pyramidprocess1} ]{%
      \includegraphics[ width=0.16\textwidth]{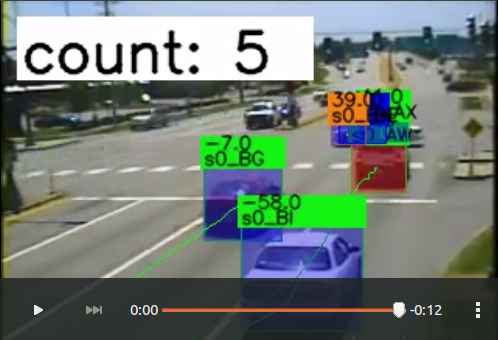}}
 \subfloat[\label{pyramidprocess2} ]{%
      \includegraphics[width=0.16\textwidth]{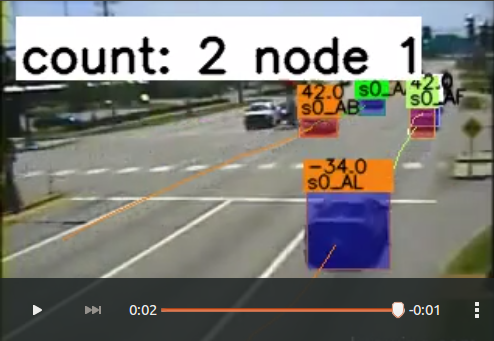}}
 \subfloat[\label{pyramidprocess3} ]{%
      \includegraphics[width=0.16\textwidth]{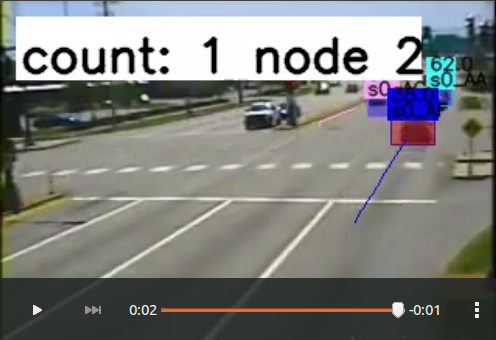}}
 \subfloat[\label{pyramidprocess4} ]{%
      \includegraphics[width=0.16\textwidth]{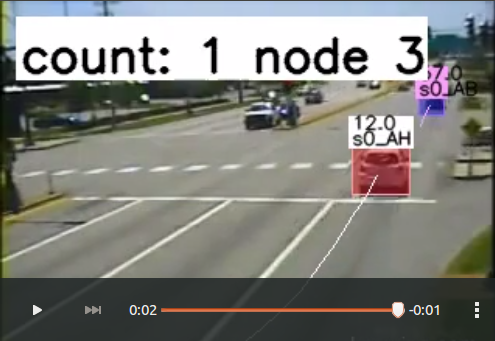}}
 \subfloat[\label{pyramidprocess5} ]{%
      \includegraphics[width=0.16\textwidth]{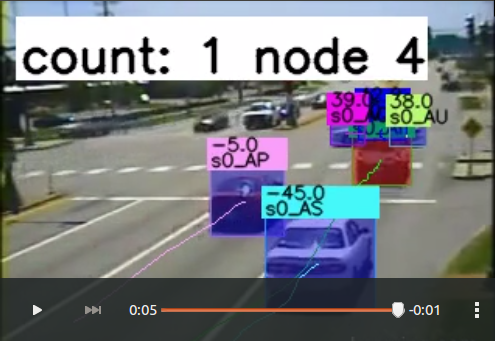}}
\caption{\label{workflow}Experiment results. (a) Street configuration; (b) entire video processed; (c),(d),(e) and (f) video processed by chunks}
\label{fig:experiment}
\end{figure*}

The filtering algorithm uses the initial street configuration (see figure \ref{fig:experiment}-a ) to define what vehicles to track, avoiding to follow the vehicles out of the select street and the vehicles that are part of other chunks. To identify the vehicles that are in more than one chunk we consider the direction of the vehicle and the street. Then, we examine if the vehicle was detected before and after the red line or if the vehicle only was detected after the line some few frames ago. If one of these conditions is satisfied, the vehicle is considered into the chunk, in any other case it is filtered using our heuristic algorithm. In the  experiment of the figure \ref{fig:experiment}, we counted the number of vehicles that pass by a specific place (red line) using two different ways: in a first approach we processed the whole video in only one Hadoop DadaNode (see figure \ref{fig:experiment}-b), a second experiment consisted on dividing the video into 4 chunks (see figures \ref{fig:experiment}-c to \ref{fig:experiment}-f) and processing each of them on a different data node. We clearly show that the final result (i.e., the vehicle frequency count) for both experiments was the same, revealing the accuracy of our approach.

\vspace{10mm}

\section*{References}

\medskip

\small

[1] Wei, Y.,\ Song, N.,\ Ke, L.,\ Chang, M. C.\ and Lyu, S.\ (2017) Street object detection/tracking for AI city traffic analysis. In 2017 IEEE SmartWorld, Ubiquitous Intelligence \& Computing Advanced \& Trusted Computed, Scalable Computing \& Communications, Cloud \& Big Data Computing, Internet of People and Smart City Innovation (SmartWorld/SCALCOM/UIC/ATC/CBDCom/IOP/SCI)\ (pp. 1-5). IEEE.

[2] Triguero, I.,\ Figueredo, G. P.,\ Mesgarpour, M.,\ Garibaldi, J. M.\ and John, R. I.\ (2017). Vehicle incident hot spots identification: An approach for big data.\ In 2017 IEEE Trustcom/BigDataSE/ICESS (pp. 901-908). IEEE.

[3] Rathore, M. M.,\ Son, H.,\ Ahmad, A.\ and Paul, A.\ (2018) Real-time video processing for traffic control in smart city using Hadoop ecosystem with GPUs,  Soft Computing {\bf 22}(5): 1533-1544

[4] Natarajan, V. A.,\ Jothilakshmi, S.\ and Gudivada, V. N.\ (2015) Scalable traffic video analytics using hadoop MapReduce, {\it ALLDATA} p 18.

[5] Yoon, I.,\ Yi, S.,\ Oh, C.,\ Jung, H.\ and Yi, Y.\ (2018) Distributed Video Decoding on Hadoop, IEICE TRANSACTIONS on Information and Systems {\bf 101}(12):2933-2941.

\end{document}